\documentclass[11pt]{article}
\pdfoutput=1
\usepackage{coling2016}
\usepackage{times}
\usepackage[T1]{fontenc}
\usepackage{url}
\usepackage{latexsym}
\usepackage{mathptmx}

\usepackage{graphicx}
\usepackage{multirow}
\usepackage{amsmath,amssymb,amstext,amsthm}
\usepackage[encapsulated]{CJK}
\usepackage{caption}
\usepackage{subcaption}

\usepackage[utf8]{inputenc} 
\usepackage{hyperref}       
\usepackage{booktabs}       
\usepackage{amsfonts}       
\usepackage{nicefrac}       
\usepackage{microtype}      
\usepackage{CJKutf8}
\usepackage{listings}

\title{Neural Contextual Conversation Learning with \\ Labeled Question-Answering Pairs}

\author{
  Kun Xiong, Anqi Cui, Zefeng Zhang \\
  RSVP Technologies Inc. \\
  Waterloo, ON, Canada \\
  \texttt{\{kun, caq, zzhang\}@rsvptech.ca} \\
  \And
  Ming Li \\
  University of Waterloo \\
  Waterloo, ON, Canada \\
  \texttt{mli@uwaterloo.ca} \\
}

\date{\today}

\begin{document}

\setlength{\pdfpagewidth}{210mm}
\setlength{\pdfpageheight}{297mm}

\maketitle

\begin{abstract}
Neural conversational models tend to produce generic or safe responses in different contexts, e.g., reply \textit{``Of course''} to narrative statements or \textit{``I don't know''} to questions.  In this paper, we propose an end-to-end approach to avoid such problem in neural generative models. Additional memory mechanisms have been introduced to standard sequence-to-sequence (seq2seq) models, so that context can be considered while generating sentences. Three seq2seq models, which memorize a fix-sized contextual vector from hidden input, hidden input/output and a gated contextual attention structure respectively, have been trained and tested on a dataset of labeled question-answering pairs in Chinese. The model with contextual attention outperforms others including the state-of-the-art seq2seq models on perplexity test. The novel contextual model generates diverse and robust responses, and is able to carry out conversations on a wide range of topics appropriately.
\end{abstract}

\section{Introduction}
A conversational dialogue model generates an appropriate response based on contextual 
information (e.g., circumstance, location, time, chatting history) and a conversational 
stimulus (i.e. utterance here). Many studies have attempted to create dialogue 
models by learning from large datasets, e.g., Twitter or movie subtitles. Data-driven 
approaches of statistical machine translation~\cite{ritter2011data} and neural 
sequence-to-sequence (seq2seq) generation~\cite{Vinyals2015A} have been adapted to 
generate conversational responses. Their major challenges are context-sensitivity, 
scalability and robustness.

\begin{figure}[ht]
\begin{center}
    \includegraphics[width=0.8\textwidth]{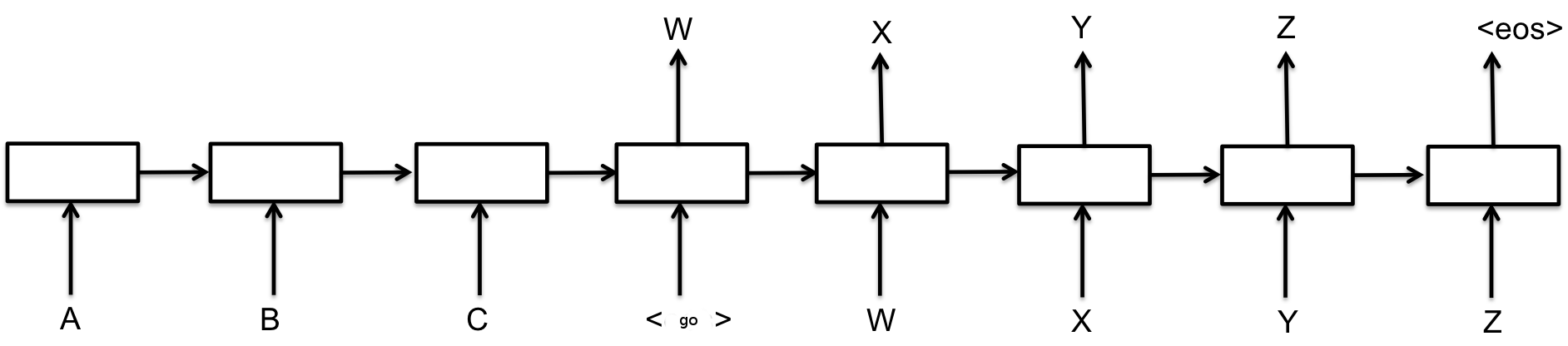}
    \caption{Basic seq2seq model of \protect\newcite{Sutskever2014Sequence}.}
    \label{fig:1}
\end{center}
\end{figure}

The great successes of recent neural language 
models~\cite{bengio2006neural,schwenk2007continuous,mnih2007three,le2011structured} 
inspired the studies of neural seq2seq learning. A significant work by 
\newcite{Sutskever2014Sequence} suggests using two recurrent neural networks (RNNs) 
to map sequences with different lengths (Figure~\ref{fig:1}). It builds 
an end-to-end machine translation model from English to French without any sophisticated 
feature engineering, in which a model is used to encode source sentences into 
fixed-length vectors, and another to generate target sentences according to the 
vectors. \newcite{bahdanau2014neural} introduced an attention mechanism on a bidirectional 
RNN-encoder and produced the state-of-the-art machine translation results.  These 
works provide a clear guideline for the subsequent seq2seq studies.  \newcite{Vinyals2015A} 
trains an end-to-end conversational system using the same vanilla seq2seq model.  
It generates related responses, but they tend to be generic, e.g., \textit{``Of course''} 
or \textit{``I don't know''}.

Recent works introduce various 
approaches~\cite{serban2016building,sordoni2015neural,li2015diversity,yao2015attention} 
to avoid such problems, and gain remarkable improvements by either encoding previous 
utterance as additional inputs or optimizing on a mutual-information function instead 
of cross-entropy.  However, they do not specify particular memory mechanism to memorize 
context and do not come to any conclusion about computing efficiency of contextual 
information.  Human conversation is smooth, because we are able to identify latent 
topics of chatting in different environments and thus provide adaptive responses.  
To simulate that, we design a conversational process that identifies the change 
of latent topics.  We find that such additional contextual information is helpful 
for seq2seq model to generate domain-adaptive responses and is effective on learning 
long-span dependencies. 



In this paper, our neural network is first trained on a community question-answering (cQA) 
dataset, and then is trained continuously on another conversation dataset. A convolutional 
neural network (CNN) has been used to extract text features and to infer latent 
topics of utterance.  A standard long short-term memory (LSTM) architecture is 
applied to process the source sentence, and another contextual LSTM is used to compute 
the target sentence.  The CNN-encoder and the RNN-encoder are both connected to 
the RNN-decoder.  They together estimate a conditional probability distribution 
of output sentences, given input sentences and contextual labels.  Our main contributions 
are: (1) We improve the conversational response generation by inventing the contextual training; 
(2) Our conversation learning is an end-to-end approach without feature engineering nor external knowledge; (3) We 
create three different mechanisms that memorizes contextual information and evaluate them.

\section{Related Work}

Natural language conversation has been a popular topic in the field of natural language 
processing.  In different practical scenarios, conversations are reduced to some traditional 
NLP tasks, e.g., question-answering, information retrieval and dialogue management.  
Recently, neural network-based generative models have been applied to generate responses 
conversationally, since these models capture deeper semantic and contextual relevancy.  
With the help of user-generated contents such as Twitter and cQA 
websites, these conversational corpora has become good resources as large-scaled training 
data~\cite{sordoni2015neural,serban2016building}.  Following this strategy, researchers 
have started to solve more challenging tasks, such as dynamic contexts~\cite{sordoni2015neural}, 
discourse structures with attention and intention~\cite{yao2015attention}, and response 
diversity by maximizing mutual information~\cite{li2015diversity}.

The evaluation of conversations, i.e., to judge if a conversation is ``good'', still lacks 
of good measurement metrics.  Ideally, a good conversation should be not only coherent, but 
also informative.  \newcite{shang2016overview} has proposed four criteria to judge the 
appropriateness of responses: Coherent, topically relevant, context-independent and 
non-repetitive.  However, this task focuses on single-round responses; it does not 
consider the contexts thus is different from our goal.  Moreover, it is difficult to quantify 
these criteria automatically with computational algorithms.

In the field of machine translation, the bilingual evaluation understudy (BLEU) algorithm has 
been traditionally used to evaluate the quality of translated texts.  This measurement 
captures the language model from the word level, and achieves a high correlation with 
human judgments.  However, in recent years, the perplexity measurement shows a better 
performance on judging languages in open domains~\cite{DBLP:journals/corr/LuongSLVZ14}.  
It is widely used to evaluate neural network-based language learning tasks.  Note that 
the scale of perplexity scores of tasks in different languages differ greatly.  
For example, an RNN encoder-decoder model for English-to-French translation has a perplexity 
score of $45.8$~\cite{Cho2014Learning}, while an attention-free German to English translation 
model has a score of $12.5$, and $8.3$ in reverse~\cite{DBLP:journals/corr/LuongLSVK15}.  
Moreover, for English to French it could be even lower at $5.8$~\cite{Sutskever2014Sequence}. 
This is natural since the complexity of languages differ from each other.  Nevertheless, 
the relative differences of models on the same task could still reflect the improvement.  
For example, \newcite{Vinyals2015A} has proved the effectiveness of an seq2seq recurrent model 
over the traditional $n$-gram based methods: It shows the perplexity scores of $8$ and $17$ 
for the seq2seq model, compared with $18$ and $28$ for the $n$-gram model, on a close-domain 
of IT helpdesk troubleshooting and an open domain of movie conversations, respectively.  
In our experiments (in Chinese), the absolute perplexity scores tend to be 
higher; but similarly, the comparison could demonstrate the effectiveness 
of our model with relatively lower scores.

\section{Contextual Models}
\begin{figure}[ht]
\begin{center}
    \includegraphics[width=0.4\textwidth]{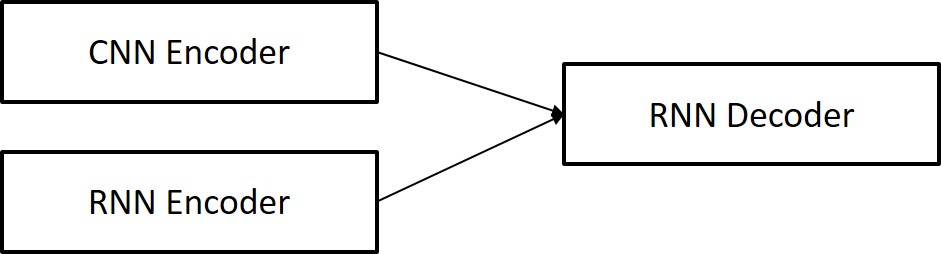}
    \caption{The Context-LSTM models architecture.}
    \label{fig:architecture}
\end{center}
\end{figure}

Our contextual seq2seq model follows the architecture of Figure~\ref{fig:architecture}.  
It takes advantage of an additional CNN-encoder that memorizes useful information 
from the context, thus it achieves better performance of sentence generation.

\subsection{CNN Contextual Encoder}
\begin{figure}[ht]
\centering
    \includegraphics[width=0.95\textwidth]{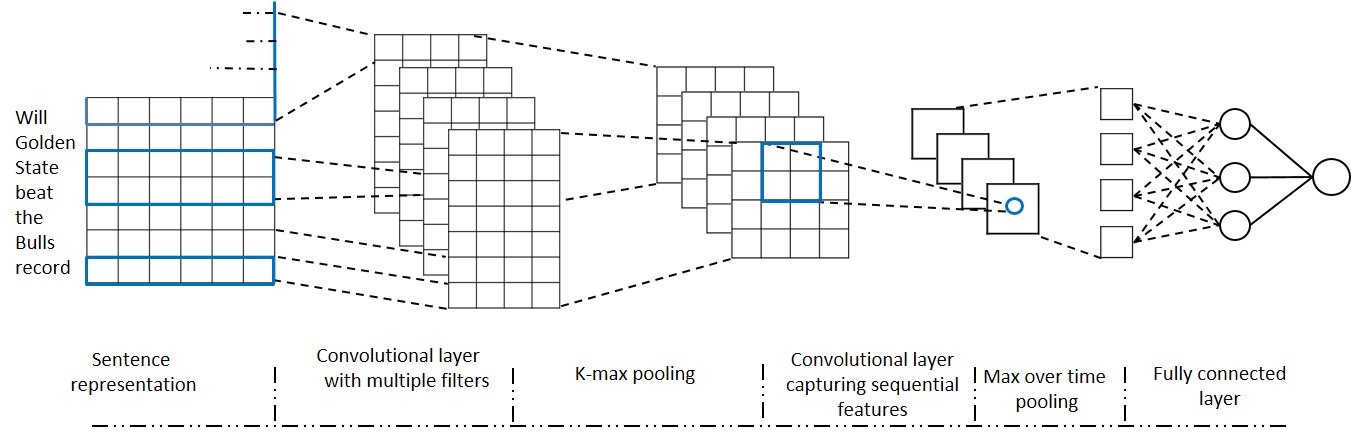}
    \caption{Structure of the CNN contextual encoder.}
    \label{fig:syntacticcnn}
\end{figure}

Instead of depending on external topic models \cite{Mikolov2012Context,ghosh2016contextual}, we have a CNN topic inferencer to learn topic distribution from questions and their labels\footnote{Please refer to Section \ref{sec:dataset} for the explanation of the labels.}.  
We build the CNN based on a simple but effective sentence classifier by \newcite{kim2014convolutional}, and add a dynamic $k$-max pooling layer and choose different hyperparameters to better fit the Chinese character-level learning (Figure~\ref{fig:syntacticcnn}).  
The widths of first-layer filters are fixed to the embedding size as suggested by \newcite{kim2014convolutional}.  Meanwhile, 
their heights are set from $1$ to $4$, as over $99\%$ of the Chinese words consist of no more than four characters in the cQA dataset.  The CNN firstly extracts basic word features, then computes syntactic features and infers semantic representation at the succeeding layers. 

Instead of producing classification results, the CNN model generates a fix-sized vector representing a probability distribution in the topic space.  We infer the topic vector from a concatenated utterance of historical conversation in the following equation:
\[ c_{\tau} = g(X_{\tau} \sqcap X_{\tau-1} \sqcap ... ) \]
where $c_{\tau}$ and $X_{\tau}$ indicates topic representation and character sequence of utterance at round $\tau$. In this setting, it is flexible to compute various length of context but does not increase gradient computation, in comparison to an RNN contextual encoder.

\subsection{RNN Contextual Decoder}
A basic RNN computes output $y_t$ from an input $x_t$ in sequence $x_1, x_2, \ldots, x_T$ 
at time $t$ as following:
\[ h_t = f(W_\textrm{hx}x_t + W_\textrm{hh}h_{t-1}) \]
\[ y_t = W_\textrm{yh}h_t \]
We apply the encoder-decoder seq2seq approach~\cite{Sutskever2014Sequence} on conversation learning. 
The model estimates the conditional probability $p(y_1,\ldots,y_{T'}|x_1,\ldots,x_T)$ 
of the source sequence $(x_1, \ldots, x_T)$ and the target sequence $(y_1, \ldots, y_{T'})$.  
To compute this probability, the LSTM-encoder computes the fix-sized representation 
$v$ from the source, and then the decoder computes the target sequence by:
\[ p(y_1,...,y_{T'}|x_1,...,x_T)=\prod_{t=1}^{T'}p(y_t|v,y_1,...,y_{t-1}) \]
In this paper, we add another CNN-encoder to the seq2seq architecture.  The RNN 
decoder depends not only on an RNN-encoder but also on this CNN-encoder.  As mentioned 
previously, the CNN produces a contextual vector $c$ from 
the question. Our contextual 
seq2seq model estimates a slightly different conditional probability:
\[ p(y_1,...,y_{T'}|x_1,...,x_T)=\prod_{t=1}^{T'}p(y_t|v,c,y_1,...,y_{t-1})\]
We build three types of contextual encoder-decoder models with different structures 
to memorize the contextual information.  The models share a same structured CNN-encoder 
and RNN-encoder, but have different contextual RNN decoders.

\subsection{Context-In Model}
The idea of the first model is to let the
LSTM memorize the context with language together.  
The LSTM uses a forget gate $f_t$ and an input gate $i_t$ to update its 
memory.  With the contextual vectors, a contextual-LSTM (CLSTM)~\cite{ghosh2016contextual} 
is able to compute the gates with contexts, by:
\begin{eqnarray*}
f_t &=& \sigma(W_f[h_{t-1}, x_t] +b_f + \boldsymbol{W_\textrm{cx}c})\\
i_t &=& \sigma(W_i[h_{t-1}, x_t] +b_i+ \boldsymbol{W_\textrm{cx}c})\\
C_t &=& f_t * C_{t-1} + i_t * \tanh(W_C[h_{t-1},x] + b_C+ \boldsymbol{W_\textrm{cx}c})\\
o_t &=& \sigma(W_0[h_{t-1}, x_t] +b_o+ \boldsymbol{W_\textrm{cx}c})\\
h_t &=& o_t * \tanh(c_t)
\end{eqnarray*}
where $\boldsymbol{c}$ is the contextual vector and $W_\textrm{cx}$ is the weight of 
the vector. 

\begin{figure}
  \centering
   \begin{subfigure}[b]{0.3\textwidth}
     \includegraphics[width=0.9\textwidth]{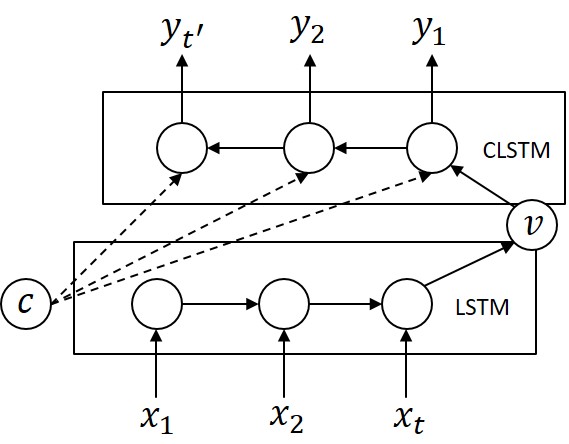}
     \caption{Context-In.}
     \label{fig:models:contextin}
   \end{subfigure}
   \hspace{0.005\textwidth}
   \begin{subfigure}[b]{0.3\textwidth}
     \includegraphics[width=0.9\textwidth]{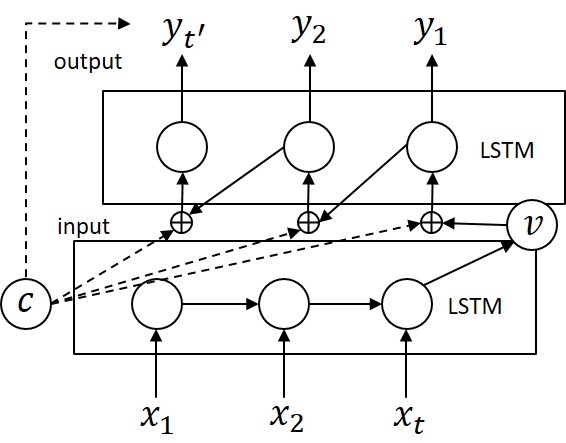}
     \caption{Context-IO.}
     \label{fig:models:contextio}
   \end{subfigure}
   \hspace{0.005\textwidth}
   \begin{subfigure}[b]{0.3\textwidth}
     \includegraphics[width=0.9\textwidth]{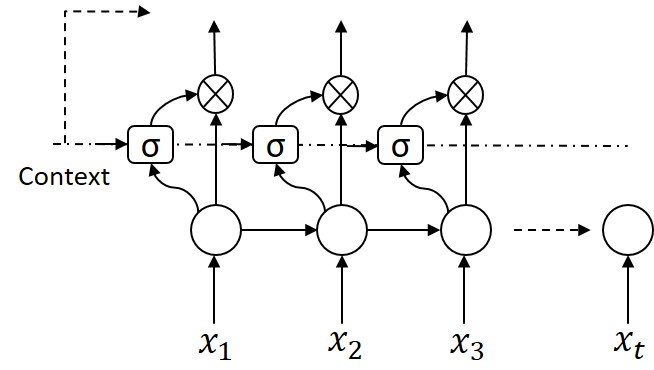}
     \caption{Context-Attn.}
     \label{fig:models:contextattn}
   \end{subfigure}
   \caption{Three contextual models.}
   \label{fig:contextmodels}
 \end{figure}

Hence the Context-In model is built as shown in Figure~\ref{fig:models:contextin}.


\subsection{Context-IO model}
Inspired by \newcite{Mikolov2012Context}, the decoder network observes context both 
at the hidden input layer and the output layer. Instead of improving a basic 
RNN language model (in the original paper), we apply such settings in the LSTM decoder of a standard seq2seq model to build the Context-IO model (shown in Figure~\ref{fig:models:contextio}):
\begin{eqnarray*}
s(t) &=& \textrm{LSTM}(W_x x_{t-1} + \boldsymbol{W_\textrm{cx}c} \cdot C_{t-1})\\
y(t) &=& \textrm{softmax}(W_y y_{t-1} + \boldsymbol{W'_\textrm{cx}c})
\end{eqnarray*}

\subsection{Context-Attention model}
The previous models apply the context computation intuitively.  An advanced strategy 
is to involve contextual vectors in the attention computation.  The Context-Attention 
(Context-Attn) model 
applies a novel contextual attention structure shown in Figure~\ref{fig:models:contextattn}.  
It uses gates to update the attention inputs.  Each gate is computed by 
the source output $h_t$ and the contextual vector $c$ by:
\[ g_t = \sigma(W^c_t\cdot c + W^h_t \cdot h_t + b_c)\]

The updated source outputs are sent to a one-layer CNN to compute the attention vector.  
The attention vector is computed at each target input of its RNN-decoder.

\section{Experiments}

\subsection{The Topic-Aware Dataset}
\label{sec:dataset}
In cQA websites, users post questions under specific categories.  After a question 
is posted, other users will then answer it, just as providing appropriate responses.  
Considering the question category as the context, these question-answer (QA) pairs 
can be used as good sources of topic-aware sentences and responses. We list a few 
examples in Table~\ref{tab:qapairexample}.



\begin{CJK*}{UTF8}{gbsn}
\begin{table}[ht]
\caption{Examples of the cQA data.}
\label{tab:qapairexample}
\centering
\small
\begin{tabular}{|c|l|}
\hline

Category & \multicolumn{1}{c|}{Question-Answer Pair} \\

\hline

\multirow{4}{*}{Movie} & Q: 2015年有成龙的电影嘛 \\
  & Are there any movies by Jackie Chan in 2015? \\
    \cline{2-2}
  & A: 有两部，《天将雄师》，另外一部是好莱坞电影《跨境追捕》 \\
  &  There are two of them: {\it Dragon Blade} and the other one {\it Skiptrace} from Hollywood. \\
\hline

\multirow{4}{*}{Sports} & Q: 勒布朗詹姆斯明年还能进决赛吗 \\
  & Will LeBron James be in the NBA final next year? \\
    \cline{2-2}
  & A: 不好说，得看乐福和欧文能否康复 \\
  & It depends on the recovery of Love and Kyrie Irving. \\
\hline
\multirow{5}{*}{Science} & Q: 天空为什么是蓝色的 \\
  & Why is the sky blue? \\
    \cline{2-2}
  & A: 晴天天空呈现蓝色是因为空气粒子散射蓝光多于红光 \\
  & A clear cloudless sky appears to be blue, because the air molecules scatter blue light \\ 
  & from the sun more than red light. \\

\hline
\end{tabular}
\end{table}
\end{CJK*}
We have collected over $200$ million QA pairs from two biggest commercial 
cQA websites in China: Baidu Zhidao and Sogou 
Wenwen\footnote{Baidu Zhidao: http://zhidao.baidu.com/, Sogou Wenwen: http://wenwen.sogou.com/}.  
In these websites, 
the categories are organized in a hierarchical structure; users may choose 
a category in any level.  To reduce the errors introduced by users,
we manually select $40$ categories according to three aspects: 
popularity, overlapping with other categories, and ambiguity of the 
category definition.  For example, the categories literature, music, movie, 
and medical are selected, but the categories entertainment, dating, 
and neurology are not selected. We have also merged the category trees from 
different websites before the selection.

Some of the questions do not have good answers for whatever reasons.  
Otherwise, at least one of the answers is marked as the best answer 
by human.
This mark is a good indicator of the quality of questions and answers.  
Therefore, we select QA pairs that have at least one best answer within the 
$40$ categories, resulting in ten million in total.

\subsection{Conversational Dataset}
The conversation dataset is acquired from two popular forum websites: Baidu 
Tieba and Douban\footnote{Baidu Tieba: https://tieba.baidu.com/, Douban: https://www.douban.com/}.  
We have collected approx. 100 million open-domain posts with comments.  The data 
is then cleaned and organized to sets of independent dialogues, in which each dialogue 
contains multiple turns of chats between two people alternately (Table~\ref{tab:convexample}).  
For training, each adjacent chat -- a pair of sentences -- is treated as the source 
and target utterance, while the sentences before them are used to infer contextual topics.

\begin{CJK*}{UTF8}{gbsn}
\begin{table}[ht]
\caption{Examples of the conversation data.}
\label{tab:convexample}
\centering
\small
\begin{tabular}{|c|l|}
\hline
Role & \multicolumn{1}{c|}{Utterance}\\
\hline
\multirow{2}{*}{Alice} & 好想要个数学大神带我装逼带我飞 \\
      & I really want a master of mathematics to lead me forward. \\
  \cline{1-2}
\multirow{2}{*}{Bob} & 这种学长应该都在被各种考试折磨中... \\
    & They might be suffering from all kinds of examinations. \\
  \cline{1-2}
\multirow{2}{*}{Alice} & 不一定哦\, // 也许人家是天才 \\
      & It is hard to say. // There must be some geniuses. \\
  \cline{1-2}
\multirow{2}{*}{Bob} & 为了理想的职业只能奋斗啊... \\
    & But they have to work hard for their dreams too. \\
\hline
\end{tabular}
\end{table}
\end{CJK*}

\subsection{Experiment Settings and Results}

Our proposed contextual models rely on a CNN-encoder, pre-trained on questions and 
their category labels.  Given an utterance as the input, the CNN-encoder turns it 
into a topic vector of size $40$.  To prove its efficiency, cross validations of 
label classification is conducted on the Chinese dataset.  The model of 
\newcite{kim2014convolutional} produces an accuracy of $75.8\%$ trained on the same 
dataset, by contrast, $77.2\%$ is reported by our CNN model.  In our experiments, 
the topic vectors with a fixed size are computed on the previous utterance and the 
current utterance.  It is used as the contextual information in the succeeding experiments.


We evaluate two types of the encoder-decoder networks, two baseline models, and three 
contextual models.  The baseline models include \newcite{Sutskever2014Sequence} and 
\newcite{bahdanau2014neural}, using the same settings in the  original papers.  
They all have the same RNN-encoder, implemented with a $3$-layer LSTM, sized $1,000$.  
The dropout technique is applied in each LSTM cell and output layers.  All these 
models are trained on the cQA dataset and then on the conversation dataset.  For 
the contextual models, contextual vectors are computed by current questions when 
training on the cQA dataset and computed by concatenated utterances of previous 
and current chats while training on the conversation dataset.  We apply the Adam 
optimizer~\cite{kingma2014adam} on training with GPU accelerators.  For testing, 
we randomly select $2,000$ pairs of utterances from both datasets, exclusively from 
the training set.


\begin{table}[ht]
\caption{Perplexities of models on sentences of different lengths.}
\label{tab:modelperplexity}
\centering
\begin{tabular}{|l|c|c|}
\hline
  \multicolumn{1}{|c|}{Models} & Short Sentences (length $<20$) & Long Sentences (length $>30$) \\
\hline
  \newcite{Sutskever2014Sequence} & $15.50$ & $33.46$ \\
  \newcite{bahdanau2014neural} & $14.10$ & $28.12$ \\
  Context-In & $14.20$   & $30.50$ \\
  Context-IO & $14.10$   & $29.50$ \\
  Context-Attn & $13.75$  & $27.00$ \\
\hline
\end{tabular}
\end{table}

In these experiments, we learn conversation on the character level too.  The performances 
are evaluated by perplexity.  However, the perplexity differ greatly between short 
sentences and long sentences, hence we divide them into two groups for a clearer 
comparison.  Generally, shorter sentences generated by the models are better -- with 
smaller perplexity -- than longer sentences.  It is most likely that the gradients 
are vanishing in long recursions, though LSTM is already applied.  

From Table~\ref{tab:modelperplexity}, we see that the Context-Attn model achieves 
overall the best perplexity. 
It works surprisingly well for the conversation learning task, since the additional memory structure creates local connections from each source LSTM to each target LSTM. The attention mechanism is an independent process from RNN, thus it reduce the long-span learning problem by establishing direct dependencies.
Models with context settings achieve smaller perplexity scores than the 
vanilla LSTM model~\cite{Sutskever2014Sequence}, since the additional memory of context is static. While decoding target sequences, it helps to further avoid the gradient vanishing problem by feeding the additional information to decoder RNN at each time.
It explains why combing attention and context in Context-Attn gains better performance.

However, perplexity only indicates how well a model predicts a target sequence.  
Low perplexity does not imply good quality of generating conversation or answering 
questions.  Therefore, we apply beam search to sample the most likely sequences on 
the RNN-decoder.  Based on human evaluations, we find that the Context-Attn model 
performs slightly better than others and it is able to overcome the 
generic-generating problem.  The following examples demonstrate the conversation 
quality of Context-Attn:

\begin{CJK*}{UTF8}{gbsn}

{\small
{\bf Chatting (Sports)}

\begin{tabular}{rll}
  {\it Human} & 你好 & Hi! \\
  {\it Machine} & 你好，你好  & Hello, hello. \\
  {\it Human} & 你是谁 & Who are you? \\
  {\it Machine} & 我是我 & I am myself. \\
  {\it Human} & 你喜欢体育运动吗 & Do you like sports? \\
  {\it Machine} & 恩，我喜欢健身 & Yes, I love fitness. \\
  {\it Human} & 平常都去哪家健身房 & Where do you workout usually? \\
  {\it Machine} & 浩沙健身房不错 & I'd like to go to Haosha Fitness. \\
  {\it Human} & 我不太喜欢健身，太累了。你喜欢篮球吗 & I don't really like workout, it's too heavy. Do you like basketball? \\
  {\it Machine} & 我喜欢篮球 & I like basketball. \\
  {\it Human} & 你觉得库里和詹姆斯谁能夺得总冠军 & Who do you think will win the final champion, Curry or James? \\
  {\it Machine} & 我觉得是詹姆斯 & I believe James will win. \\
\end{tabular}
}

{\small
{\bf Chatting (Movie)}

\begin{tabular}{rll}
  {\it Human} &  最近上映了好多电影  & Many movies are released recently. \\
  {\it Machine} & 什么电影  & What movies? \\
  {\it Human} &  盗梦空间蛮不错的  & \textit{Inception} is pretty impressive. \\
  {\it Machine} & 当然好看了  & Of course, it is good. \\
  {\it Human} & 你最喜欢哪部电影  & Which movie do you like best? \\
  {\it Machine} & 我最喜欢白蛇传  & I love \textit{Legend of the White Snake}. \\
  {\it Human} & 白蛇传是谁演的  & Who was in the movie? \\
  {\it Machine} & 霍建华演的*  & Wallace Huo played. \\
  {\it Human} & 为什么许仙不能和白素贞在一起 & Why could't Xu Xian and Bai Suzhen be together? \\
  {\it Machine} & 因为许仙也是女人*  & Because Xu Xian was a woman too. \\
  {\it Human} & 好吧，你赢了  & OK, you win. \\
  {\it Machine} &  哈哈  & Haha! \\
\end{tabular}
}

{\small
{\bf Question Answering}

\begin{tabular}{rll}
  {\it Human} &  美国总统是谁  & Who is the president of the United States? \\
  {\it Machine} &  奥巴马  & Obama. \\
  {\it Human} &  中国有多少人口  & How many people live in China? \\
  {\it Machine} & 14亿多吧  & About 1.4 billion. \\
  {\it Human} & 世界最高峰在哪里  & Where is the highest mountain in the world? \\
  {\it Machine} & 珠穆朗玛峰峰顶  & The peak of Mount Everest. \\
  {\it Human} & 谁发明了电灯  & Who invented the light bulb? \\
  {\it Machine} & 爱迪生发明了电灯  & Edison invented the light bulb. \\
  {\it Human} & 法国工业革命在什么时候  &  When did the industrial revolution take place in France? \\
  {\it Machine} & 文艺复兴时期*  & It was during the Renaissance period. \\
  {\it Human} & 中国的国歌是什么  & What is the Chinese national anthem? \\
  {\it Machine} &  国歌是义勇军进行曲  & The national anthem is \textit{March of the Volunteers}. \\
\end{tabular}
}

\end{CJK*}

These examples illustrate that our proposed model generates reasonable responses 
with domain-specific vocabulary, while avoid safe but general answers to some 
extent.  Even for the task of question answering, it shows the capability of 
providing (mostly) correct answers.  The reason is that the contextual attention 
structure memorizes important (or frequent) information, which is usually the 
answer to the question.  In some cases (marked with * in the examples), the answers 
are incorrect.  For example, Wallace Huo has played in neither movies 
nor TV series on the {\it Legend of the White Snake};
Xu Xian was actually a man (although in a 
TV show he was played by an actress); and the 
industrial revolution in France took place more than $300$ years after the 
Renaissance.  It indicates the memory itself works differently from a real 
question-answering mechanism.

To further demonstrate the efficiency of our contextual settings, we visualize the 
weights in original soft attention and our contextual gated attention in 
Table~\ref{tab:visualization}.  For the background of the sentence, darker color 
represents larger value of weights.  Sentences are translated to English literally 
to show the correspondence of words.  The Context-Attn model estimates a conditional 
probability distribution of responses given source sentences and context vectors.  
The additional gates in the contextual attention automatically determine which to 
augment and which to eliminate by computing contextual information.  Therefore, 
context is able to manipulate the generation process of the characters in the LSTM model.  
That explains why {\it Titanic} and {\it James} have higher weights.  The contextual 
attention helps generate domain-adaptive sentences.  It is also considered to be 
flexible and efficient, since such a gated attention works similarly to a standard 
soft attention and is able to simulate a hard attention in extreme case at the 
same time. 

\begin{CJK*}{UTF8}{gbsn}
\begin{table}[h]
    \caption{Visualization of weights in the soft attention and contextual attention.}
    \label{tab:visualization}
\begin{center}
\small
\begin{tabular}{|c|c|l|}
\hline
Context & Question & \multicolumn{1}{c|}{Answer} \\
\hline
-- & \parbox[c]{120px}{\includegraphics[width=\linewidth]{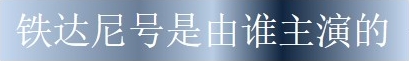}} & 有很多很多人 (There are many many actors.) \\
 & {\it Titanic} is by whom performed & \\
Movie & \parbox[c]{120px}{\includegraphics[width=\linewidth]{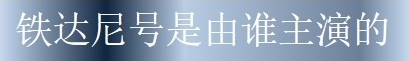}} & 莱昂纳多，迪卡普里奥 (Leonardo DiCaprio.) \\
\hline
-- & \parbox[c]{120px}{\includegraphics[width=\linewidth]{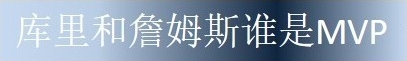}} & 那要看情况 (It depends.) \\
 & Curry and James, who is the MVP & \\
Sports & \parbox[c]{120px}{\includegraphics[width=\linewidth]{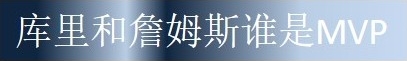}} & 答案是詹姆斯 (The answer is James.) \\
\hline
\end{tabular}
\end{center}
\end{table}
\end{CJK*}

\subsection{Robust Chatting}
While we were talking about the four criteria of 
\newcite{shang2016overview} of coherency, relevancy, context-independence
and non-repetivieness for judging a good conversation, 
at the end of this article, we wish to mention another
important dimension of conversation: robustness. 
No matter how well we can chat, if the chatting works only along a 
very narrow pathway, and the chatbot does not tolerate any variations,
the user experience would be very bad. This problem has been troubling
template-based chatting systems for being too restrictive, and
keyword-based systems for having too many false positives.

Our Context-Attn model solves this problem as a side product wonderfully,
as the following off-hand random experiment reveals.  
Italic words denote some noises in the sentences.

\begin{CJK*}{UTF8}{gbsn}
\small
\begin{tabular}{rll}
  {\it Human} & 我渴了怎么办 & I am thirsty, what do I do? \\
  {\it Machine} & 喝水  & Drink water. \\
  \multirow{4}{*}{\it Human} & (1) 我{\CJKfamily{gkai}有点}渴了怎么办 & I am {\em a bit} thirsty, what do I do? \\
   & (2) 我{\em x}有点渴了怎么办 & I {\em x} a bit thirsty, what do I do? \\
   & (3) 我{\CJKfamily{gkai}来}有点渴了怎么办 & I {\em come} a bit thirsty, what do I do? \\
   & (4) 我{\CJKfamily{gkai}中国}有点渴了怎么办 & I {\em China} a bit thirsty, what do I do? \\
  {\it Machine} & 多喝水 & Drink more water. \\
  {\it Human} & 我{\CJKfamily{gkai}北京}有点渴了怎么办{\CJKfamily{gkai}中国} & I {\em Beijing} a bit thirsty, what do I do {\em China}? \\
  {\it Machine} & 喝王老吉 & Drink Wong Lo Kat (a popular Chinese herbal tea drink). \\
  {\it Human} & {\CJKfamily{gkai}蝴蝶}我{\CJKfamily{gkai}北京}有点渴了怎么办{\CJKfamily{gkai}中国} & {\em butterfly} I {\em Beijing} a bit thirsty, what do I do {\em China}? \\
  {\it Machine} & 喝点水 & Drink a bit water. \\
  {\it Human} & {\CJKfamily{gkai}蝴蝶}我{\CJKfamily{gkai}北京}有点饿了怎么办{\CJKfamily{gkai}中国} & {\em butterfly} I {\em Beijing} a bit hungry, what do I do {\em China}? \\
  {\it Machine} & 吃东西 & Eat something. \\
  {\it Human} & 蝴蝶在北京有没有{\CJKfamily{gkai}怎么办中国} & butterfly in Beijing are there any {\em what do I do China}? \\
  {\it Machine} & 有啊，在北京 & Yes there are (butterflies), in Beijing. \\
\end{tabular}
\end{CJK*}

\section{Conclusion}
In this paper, we target on domain-adaptive and robust conversation generation by end-to-end learning without any feature-engineering.
We have introduced a CNN-encoder to infer latent topics of source sentences to 
seq2seq models and created various external memory structure for considering 
contexts; the gated attention mechanism is the most efficient mechanism to capture 
the contextual information, reflected in the generated responses. 
The Context-Attn model also outperforms traditional seq2seq models on perplexity tests.
The experiments reveal that training on the QA dataset helps conversation 
generation from two aspects: 1) Gain context-awareness from the question-label 
learning; 2) Gain additional robustness from the question-answer learning.  
Our proposed model is demonstrated to generate interesting and 
context-sensitive responses from variations of source sentences.
Our future works will be further improving the robustness and consistency.

\bibliographystyle{acl}

\bibliography{context_encdec}

\begin{thebibliography}{}

\bibitem[\protect\citename{Bahdanau \bgroup et al.\egroup
  }2014]{bahdanau2014neural}
Dzmitry Bahdanau, Kyunghyun Cho, and Yoshua Bengio.
\newblock 2014.
\newblock Neural machine translation by jointly learning to align and
  translate.
\newblock {\em arXiv preprint arXiv:1409.0473}.

\bibitem[\protect\citename{Bengio \bgroup et al.\egroup
  }2006]{bengio2006neural}
Yoshua Bengio, Holger Schwenk, Jean-S{\'e}bastien Sen{\'e}cal, Fr{\'e}deric
  Morin, and Jean-Luc Gauvain.
\newblock 2006.
\newblock Neural probabilistic language models.
\newblock In {\em Innovations in Machine Learning}, pages 137--186. Springer.

\bibitem[\protect\citename{Cho \bgroup et al.\egroup }2014]{Cho2014Learning}
Kyunghyun Cho, Bart~Van Merrienboer, Caglar Gulcehre, Dzmitry Bahdanau, Fethi
  Bougares, Holger Schwenk, and Yoshua Bengio.
\newblock 2014.
\newblock Learning phrase representations using rnn encoder-decoder for
  statistical machine translation.
\newblock {\em Eprint Arxiv}.

\bibitem[\protect\citename{Ghosh \bgroup et al.\egroup
  }2016]{ghosh2016contextual}
Shalini Ghosh, Oriol Vinyals, Brian Strope, Scott Roy, Tom Dean, and Larry
  Heck.
\newblock 2016.
\newblock Contextual lstm (clstm) models for large scale nlp tasks.
\newblock {\em arXiv preprint arXiv:1602.06291}.

\bibitem[\protect\citename{Kim}2014]{kim2014convolutional}
Yoon Kim.
\newblock 2014.
\newblock Convolutional neural networks for sentence classification.
\newblock {\em arXiv preprint arXiv:1408.5882}.

\bibitem[\protect\citename{Kingma and Ba}2014]{kingma2014adam}
Diederik Kingma and Jimmy Ba.
\newblock 2014.
\newblock Adam: A method for stochastic optimization.
\newblock {\em arXiv preprint arXiv:1412.6980}.

\bibitem[\protect\citename{Le \bgroup et al.\egroup }2011]{le2011structured}
Hai-Son Le, Ilya Oparin, Alexandre Allauzen, Jean-Luc Gauvain, and
  Fran{\c{c}}ois Yvon.
\newblock 2011.
\newblock Structured output layer neural network language model.
\newblock In {\em Acoustics, Speech and Signal Processing (ICASSP), 2011 IEEE
  International Conference on}, pages 5524--5527. IEEE.

\bibitem[\protect\citename{Li \bgroup et al.\egroup }2015]{li2015diversity}
Jiwei Li, Michel Galley, Chris Brockett, Jianfeng Gao, and Bill Dolan.
\newblock 2015.
\newblock A diversity-promoting objective function for neural conversation
  models.
\newblock {\em arXiv preprint arXiv:1510.03055}.

\bibitem[\protect\citename{Luong \bgroup et al.\egroup
  }2014]{DBLP:journals/corr/LuongSLVZ14}
Thang Luong, Ilya Sutskever, Quoc~V. Le, Oriol Vinyals, and Wojciech Zaremba.
\newblock 2014.
\newblock Addressing the rare word problem in neural machine translation.
\newblock {\em CoRR}, abs/1410.8206.

\bibitem[\protect\citename{Luong \bgroup et al.\egroup
  }2015]{DBLP:journals/corr/LuongLSVK15}
Minh{-}Thang Luong, Quoc~V. Le, Ilya Sutskever, Oriol Vinyals, and Lukasz
  Kaiser.
\newblock 2015.
\newblock Multi-task sequence to sequence learning.
\newblock {\em CoRR}, abs/1511.06114.

\bibitem[\protect\citename{Mikolov and Zweig}2012]{Mikolov2012Context}
T.~Mikolov and G.~Zweig.
\newblock 2012.
\newblock Context dependent recurrent neural network language model.
\newblock In {\em Spoken Language Technology Workshop (SLT), 2012 IEEE}, pages
  234--239.

\bibitem[\protect\citename{Mnih and Hinton}2007]{mnih2007three}
Andriy Mnih and Geoffrey Hinton.
\newblock 2007.
\newblock Three new graphical models for statistical language modelling.
\newblock In {\em Proceedings of the 24th international conference on Machine
  learning}, pages 641--648. ACM.

\bibitem[\protect\citename{Ritter \bgroup et al.\egroup }2011]{ritter2011data}
Alan Ritter, Colin Cherry, and William~B Dolan.
\newblock 2011.
\newblock Data-driven response generation in social media.
\newblock In {\em Proceedings of the conference on empirical methods in natural
  language processing}, pages 583--593. Association for Computational
  Linguistics.

\bibitem[\protect\citename{Schwenk}2007]{schwenk2007continuous}
Holger Schwenk.
\newblock 2007.
\newblock Continuous space language models.
\newblock {\em Computer Speech \& Language}, 21(3):492--518.

\bibitem[\protect\citename{Serban \bgroup et al.\egroup
  }2016]{serban2016building}
Iulian~V Serban, Alessandro Sordoni, Yoshua Bengio, Aaron Courville, and Joelle
  Pineau.
\newblock 2016.
\newblock Building end-to-end dialogue systems using generative hierarchical
  neural network models.
\newblock In {\em Proceedings of the 30th AAAI Conference on Artificial
  Intelligence (AAAI-16)}.

\bibitem[\protect\citename{Shang \bgroup et al.\egroup
  }2016]{shang2016overview}
Lifeng Shang, Tetsuya Sakai, Zhengdong Lu, Hang Li, Ryuichiro Higashinaka, and
  Yusuke Miyao.
\newblock 2016.
\newblock Overview of the ntcir-12 short text conversation task.
\newblock {\em Proceedings of the 12th NTCIR Conference on Evaluation of
  Information Access Technologies}, pages 473--484.

\bibitem[\protect\citename{Sordoni \bgroup et al.\egroup
  }2015]{sordoni2015neural}
Alessandro Sordoni, Michel Galley, Michael Auli, Chris Brockett, Yangfeng Ji,
  Margaret Mitchell, Jian-Yun Nie, Jianfeng Gao, and Bill Dolan.
\newblock 2015.
\newblock A neural network approach to context-sensitive generation of
  conversational responses.
\newblock {\em arXiv preprint arXiv:1506.06714}.

\bibitem[\protect\citename{Sutskever \bgroup et al.\egroup
  }2014]{Sutskever2014Sequence}
Ilya Sutskever, Oriol Vinyals, Quoc~V. Le, Ilya Sutskever, Oriol Vinyals, and
  Quoc~V. Le.
\newblock 2014.
\newblock Sequence to sequence learning with neural networks.
\newblock {\em Advances in Neural Information Processing Systems},
  4:3104--3112.

\bibitem[\protect\citename{Vinyals and Le}2015]{Vinyals2015A}
Oriol Vinyals and Quoc Le.
\newblock 2015.
\newblock A neural conversational model.
\newblock {\em Computer Science}.

\bibitem[\protect\citename{Yao \bgroup et al.\egroup }2015]{yao2015attention}
Kaisheng Yao, Geoffrey Zweig, and Baolin Peng.
\newblock 2015.
\newblock Attention with intention for a neural network conversation model.
\newblock {\em arXiv preprint arXiv:1510.08565}.

\end{thebibliography}


\end{document}